# NEBench R1: a Reasoning Benchmark for Evaluating Nuclear Domain Competence in Large Language Models

Liyan Wang[a*], Longze Li[a], Mohamed Hassan[a], Isshu Lee[b], Dajie Sun[c], Ruimin Feng[d], Aleksandar Vakanski[c]

[a] *Department of Computer Science, University of Idaho, Moscow, ID 84843, USA*
[b] *Department of Material Science, Oregon State University, Corvallis, OR 97331, USA*
[c] *Department of Nuclear Engineering and Industrial Management, Idaho Falls, ID 83401, USA*
[d] *Department of Civil and Environmental Engineering, University of Idaho, Moscow, ID 84843, USA*

**Corresponding author*
*Email address: iwang@uidaho.edu*

**Abstract:** Large language models (LLMs) are increasingly being explored for knowledge-intensive and safety-critical applications in nuclear engineering. However, existing nuclear-domain benchmarks provide limited content coverage or focus primarily on answer accuracy without systematically evaluating the quality of the underlying reasoning. This study introduces NEBench R1, a reasoning-oriented benchmark comprising 7,546 unique multiple-choice questions extracted from U.S. Nuclear Regulatory Commission (NRC) reactor operator licensing examinations covering 18 nuclear facilities. Each question is classified according to the NRC Knowledge and Abilities catalogs and accompanied by an official explanation serving as a reference for reasoning evaluation. Six LLMs were evaluated, with overall accuracy ranging from 34.0% to 73.9% across three distinct performance tiers. To assess reasoning quality, an LLM-based judge evaluated correctly answered responses using a multidimensional rubric. Although responses generally demonstrated strong problem comprehension, they scored lower in factual and procedural accuracy and distractor analysis, indicating that correct answers were not necessarily supported by technically sound justifications. Response clarity was also weakly indicative of technical quality. Furthermore, longer reasoning traces were associated with lower accuracy in both frontier reasoning models, suggesting that reasoning length may reflect question difficulty rather than improved problem-solving performance. Reasoning-token consumption may therefore provide an additional signal for identifying potentially unreliable responses. NEBench R1 establishes a benchmark for evaluating both answer accuracy and reasoning quality, providing a foundation for assessing the domain competence and reliability of LLMs in nuclear engineering.

## 1. Introduction

Large language models (LLMs) are increasingly being explored as tools to support experts in safety-critical, knowledge-intensive domains, including nuclear engineering. In nuclear education and training, LLMs can provide answers to theoretical questions involving nuclear physics and thermodynamics. In nuclear digital twins, LLMs can be integrated with simulation models and digital twin systems to support system monitoring and decision-making. Other emerging applications include the automated review of technical specifications and operating procedures, incident report analysis, and reactor operator assistance [1]. However, evaluating LLM competence in nuclear engineering presents substantial challenges. Beyond factual knowledge retrieval, nuclear applications require accurate technical reasoning across interconnected disciplines, including nuclear physics, reactor theory, thermodynamics, and instrumentation and control, as well as the ability to interpret operating procedures and reason about safety-critical scenarios. Consequently, answer accuracy alone may provide an incomplete assessment of whether an LLM possesses the domain knowledge and reasoning capabilities required for these applications.

Despite growing interest in evaluating LLMs in the nuclear domain, dedicated benchmarks for nuclear engineering remain limited. NuclearQAv2 [2] comprises approximately 1,240 Boolean, numerical, and verbal question-answer pairs, primarily derived from a single textbook. NucBench [1] offers broader coverage, comprising 4,292 multiple-choice questions sourced from Nuclear Regulatory Commission (NRC) reactor operator licensing examinations. However, its evaluation primarily emphasizes answer-selection accuracy, while explicit reasoning assessment is restricted to 108 questions from a single undergraduate course. Thus, existing benchmarks do not jointly provide comprehensive coverage of licensing examination knowledge and systematic assessment of LLM reasoning capabilities. This gap motivates the development of a benchmark that evaluates both dimensions of nuclear domain competence.

**Sample Index:** 1400
**Source:** Cooper Nuclear Station

**Question:** Which one of the following statements describes subcritical multiplication during a reactor startup at a BWR?

(A) As Keff approaches unity, a smaller change in neutron level occurs for a given change in Keff.
(B) As Keff approaches unity, less time is required to reach the equilibrium neutron level for a given change in Keff.
(C) Subcritical multiplication is the process of using source neutrons to maintain an equilibrium neutron population when Keff is less than 1.
(D) The equilibrium subcritical neutron level is dependent on the source strength and the time between successive reactivity insertions.

**Correct answer: C.**

**Explanation:** Subcritical multiplication is the process of utilizing source neutrons and fuel to maintain a constant neutron population with Keff < 1. It is also the process by which the neutron population is maintained above source strength as a result of source neutrons in a subcritical reactor. Distractor A has the direction of the effect reversed — the change in neutron level grows larger, not smaller, as Keff approaches unity. Distractor B confuses the speed of response with the magnitude of multiplication. Distractor D incorrectly introduces time between reactivity insertions as a variable; the equilibrium level depends on source strength and Keff only.

**Fig. 1.** Example of a question, correct answer, and explanation.

In addition, although multiple-choice accuracy is widely used to evaluate LLM performance, it provides limited insight into the reasoning underlying model predictions. A model may select the correct answer through memorization, superficial pattern matching, or distractor elimination without demonstrating a sound understanding of the underlying concepts. This limitation is particularly consequential in nuclear engineering, where reliable problem-solving requires the application of domain knowledge, integration of concepts across technical disciplines, and systematic reasoning about complex engineering problems. Reasoning-oriented benchmarks seek to address this limitation by assessing not only final-answer correctness but also the coherence and validity of the reasoning used to derive answers, often through chain-of-thought analysis and rubric-based evaluation. Such evaluation approaches have been investigated in several scientific and technical domains. In biomedical research, MedQA-CS [3] evaluates clinical reasoning through realistic patient-encounter scenarios requiring multistep diagnostic judgment. GPQA [4] assesses expert-level knowledge and problem-solving capabilities in biology, chemistry, and physics. In materials science, MatSciBench [5] demonstrates the value of reasoning-oriented evaluation in identifying model limitations that may not be captured by answer accuracy alone. Despite these advances, a large-scale benchmark that systematically assesses both answer correctness and reasoning quality in nuclear engineering remains lacking.

To address this gap, we introduce NEBench R1, a nuclear engineering benchmark designed to evaluate LLM competence across both foundational nuclear engineering theory and reactor-specific operational knowledge. The benchmark comprises 7,546 unique question-answer pairs extracted from NRC reactor operator licensing examinations covering 18 nuclear power plants, including both boiling water reactors (BWRs) and pressurized water reactors (PWRs). Based on the NRC's official Knowledge and Abilities (K/A) catalogs [6,7], the examination content spans five major areas: Generic Knowledge and Abilities, Plant Systems, Emergency and Abnormal Plant Evolutions, Components, and Theory. The Theory area covers two principal domains: Reactor Theory and Thermodynamics Theory. Reactor Theory includes neutron behavior, the neutron life cycle, reactor kinetics, neutron sources, reactivity coefficients, control rods, fission product poisons, fuel depletion, burnable poisons, and reactor operational physics. Thermodynamics Theory encompasses fundamental energy concepts, steam properties, heat transfer, thermal hydraulics, core thermal limits, and brittle fracture and vessel thermal stress. By integrating these knowledge domains, NEBench R1 supports the evaluation of LLMs' understanding of both plant-specific operational requirements, including operating limits and emergency procedures, and the foundational engineering principles governing reactor behavior and operation.

In addition to question-answer pairs, each NEBench R1 question is accompanied by a detailed explanation specifying why the correct answer is valid and why each distractor is incorrect (Fig. 1). These explanations provide explicit reasoning annotations derived from authoritative examination materials, reducing reliance on costly, manual expert annotation. Furthermore, the use of official NRC examination materials establishes traceable question provenance, while the regulator-defined K/A classification provides a standardized and auditable taxonomy of nuclear engineering knowledge. The use of official examination sources also provides a basis for assessing potential data contamination through source-level traceability. These characteristics enable the evaluation of the technical validity of model-generated explanations, providing a more comprehensive assessment of LLM competence across foundational theory and operational practice.

This work makes the following contributions:

- A large-scale reasoning benchmark for nuclear engineering is introduced which comprises 7,546 unique question-answer pairs covering foundational nuclear engineering theory and reactor-specific operational knowledge. .
- A comprehensive evaluation of contemporary LLMs is conducted with six LLMs representing open-weight and closed-weight systems, dense and mixture-of-experts architectures, and reasoning-enabled and conventional inference configurations. The results establish a performance baseline and characterize the gap between current LLM capabilities and the NRC reactor operator examination standard.
- The relationship between computational resource consumption and answer accuracy in frontier reasoning models is investigated, providing an empirical analysis of inference-time reasoning dynamics on specialized nuclear engineering tasks. The results demonstrate that increased computational expenditure and extended reasoning do not necessarily translate into improved performance on domain-specific evaluations.

## 2. Data Curation

The source questions for NEBench R1 were collected from NRC reactor operator written examinations archived, along with their corresponding answer keys, in the NRC's Agencywide Documents Access and Management System (ADAMS) [8]. For each candidate reactor facility, the corresponding docket number was identified using the facility name and subsequently used to search ADAMS for documents containing the keyword "written exam" in the document title. Retrieved examination documents were screened, and only those providing detailed explanations for each answer choice were retained, as these explanations were necessary for constructing the reference reasoning annotations described above. This document selection process yielded examination materials from 18 reactor facilities, comprising 6 BWRs and 12 PWRs.

Table 1. Reactors type and NRC docket numbers.

| Type | Reactor facility (docket number) |
|---|---|
| PWR | Arkansas Nuclear 1(5000313), Arkansas Nuclear 2(5000368), Beaver Valley 1(5000334), Beaver Valley 2(5000412), Braidwood 1(5000456), Braidwood 2(5000457), Byron 1(5000454), Callaway (5000483), Catawba 2(5000414), Comanche Peak 1(5000445), Diablo Canyon 1(5000275), Diablo Canyon 2(5000323) |
| BWR | Browns Ferry 1(5000259), Browns Ferry 2(5000260), Brunswick 1(5000325), Clinton (5000461), Columbia Generating Station (5000397), Cooper (5000298) |

The retrieved examination PDFs were processed using Qwen3.6-35B-A3B [9] to extract structured information for each question, including its associated K/A reference, four answer choices, the correct answer, and explanations for both the correct and incorrect choices. Because questions were sometimes repeated or slightly modified across examination cycles, a deduplication procedure was applied, resulting in 7,546 unique question-answer pairs. Furthermore, because the examinations span more than a decade, revisions to the official NRC K/A catalogs introduced inconsistencies between K/A codes and their corresponding textual descriptions. To establish a consistent classification scheme across the dataset, each question was assigned to one of three categories based on its official K/A description: Knowledge, Ability, or Other. The original K/A codes and descriptions were retained to support alternative classification analyses and downstream applications. Of the 7,546 questions, 3,893 (51.6%) were classified as Knowledge, 2,898 (38.4%) as Ability, and 755 (10.0%) as Other. An example NEBench R1 entry (Fig. 1) is presented to illustrate the structure of the dataset.

## 3. Model Evaluation

### 3.1 Evaluated Models and Results

Six LLMs were evaluated on NEBench R1, representing different parameter scales, model architectures, and access configurations. These models were specifically selected to represent a broad cross-section of the current LLM landscape, spanning open-weight and proprietary systems, dense transformer and mixture-of-experts architectures, and standard autoregressive inference versus extended internal reasoning modes, thereby enabling systematic comparison across these dimensions. Four open-weight models were included: Qwen3.8-27B (Alibaba Group) [10], a dense model; Gemma 4.1 (Google DeepMind) [11], a dense model; Llama 3.3 70B Instruct (Meta Platforms) [12], a dense model; and Nemotron-Ultra-550B (NVIDIA) [13], a mixture-of-experts (MoE) model with 550 billion total parameters and approximately 55 billion active parameters per forward pass. Two closed-weight models with reasoning capabilities were also evaluated: Grok 4.6 (xAI) [14], which generates internal reasoning tokens before producing its final response, and Claude Opus 5.5 (Anthropic) [15], which supports extended reasoning. The parameter counts of both closed-weight models have not been publicly disclosed. Collectively, these models provide a basis for comparing LLM performance across different parameter scales, architectural designs, model access configurations, and reasoning capabilities.

All models were evaluated using a standardized prompt consisting of a question and four labeled answer choices. The prompt instructed each model to return its selected option using the format "Answer: X," where X denotes the corresponding answer-choice letter. Qwen3.8-27B and Gemma 4.1 were deployed locally on a server equipped with four Quadro RTX 8000 GPUs (192 GB GPU memory in total) and evaluated using a temperature of 0.3 and a maximum output length of 256 tokens. For both models, the full dataset was evaluated under non-reasoning inference, while reasoning-enabled inference was conducted on a randomly selected subset of 1,000 questions. Llama 3.3 70B, Nemotron-Ultra-550B, and Grok 4.6 were accessed through the OpenRouter API, with the same temperature and maximum output length applied where supported. Claude Opus 5.5 was accessed through the Anthropic API. For Grok 4.6 and Claude Opus 5.5, adaptive thinking was enabled without imposing a fixed reasoning-token budget. During the initial evaluation, the 256-token output limit resulted in truncation of all Nemotron-Ultra-550B responses. To eliminate this truncation artifact, Nemotron-Ultra-550B was re-evaluated with the maximum output length increased to 4,096 tokens. All reported results for this model are based on the revised evaluation.

The six models exhibited a pronounced hierarchy of performance, with overall accuracy spanning more than 40 percentage points between the weakest and strongest models.

Table 2. Overall evaluation results on NEBench R1.

| Model | Parameters | Knowledge | Ability | PWR | BWR | Overall Accuracy | Valid Responses/Full dataset |
|---|---|---|---|---|---|---|---|
| Llama 3.3 70B Instruct | 70B | 34.7% | 29.7% | 32.5% | 35.5% | 34.0% | 7,540 |
| Qwen3.8-27B | 27B | 33.0% | 32.1% | 33.6% | 35.6% | 34.5%<br>33.63% * | 7,357<br>1,000 |
| Gemma 4.1 | 31B | 45.2% | 35.9% | 40.5% | 42.3% | 41.3%<br>42.5% * | 7,517<br>1,000 |
| Nemotron-Ultra-550B | 550B (55B active) | 44.7% | 42.0% | 46.2% | 43.4% | 44.9% | 7,314 |
| Grok 4.6 | (Not disclosed) | 67.9% | 66.3% | 69.1% | 67.7% | 68.4% | 7,530 |
| Claude Opus 5.5 | (Not disclosed) | 71.3% | 70.3% | 74.6% | 73.2% | 73.9% | 7,544 |

* Reasoning enabled on 1000 random selected subset.

The six evaluated models exhibited three distinct performance tiers based on overall accuracy. The first tier comprised Llama 3.3 70B Instruct and Qwen3.8-27B, both achieving approximately 34% accuracy, representing an improvement of approximately 9 percentage points over random guessing. The second tier included Gemma 4.1 and Nemotron-Ultra-550B, which achieved intermediate accuracy levels but remained substantially below the top-performing models. The third tier consisted of Grok 4.6 and Claude Opus 5.5, whose accuracies exceeded those of the intermediate-tier models by more than 23 percentage points, indicating a pronounced performance gap across the evaluated systems. Beyond overall accuracy, performance differences across question categories and reactor types were relatively modest. For four of the six models, the absolute accuracy differences between knowledge-based and ability-based questions were within 2 percentage points. Similarly, accuracy differences between BWR and PWR questions generally ranged from 2 to 3 percentage points. To further examine the contribution of explicit reasoning, Qwen3.8-27B and Gemma 4.1 were evaluated under reasoning-enabled configurations using a randomly selected subset of 1,000 questions. Enabling reasoning resulted in accuracy improvements of less than one percentage point for both models.

The relatively small performance differences across K/A categories and reactor types suggest that the evaluated models exhibited broadly comparable capabilities across foundational nuclear knowledge, operational reasoning, and BWR/PWR examination content. More importantly, the marginal accuracy gains observed under reasoning-enabled configurations indicate that enabling explicit reasoning alone may be insufficient to address limitations in domain-specific knowledge. NRC licensing examinations frequently require precise knowledge of plant-specific parameters, operating procedures, actuation setpoints, and regulatory requirements, much of which cannot be reliably inferred from general engineering principles. When the necessary domain knowledge is unavailable, additional reasoning may offer limited benefits and can instead produce plausible but technically incorrect conclusions.

### 3.2 Model Failure Statistics

For five of the six evaluated models (Qwen3.8-27B, Gemma 4.1, Llama 3.3 70B, Grok 4.6, and Claude Opus 5.5), incorrect predictions were distributed approximately uniformly across answer choices A–D (Fig. 2), suggesting no substantial answer-position bias. In contrast, Nemotron-Ultra-550B exhibited a pronounced preference for option D, which accounted for 50.3% of its incorrect predictions and 42.6% of all its predictions, compared with 23.7% of the gold-standard answers assigned to this option. Notably, option D also represented 33.2% of the model's correct predictions, indicating that this preference persisted regardless of prediction correctness.

Across the evaluated questions, 576 (7.9%) were answered incorrectly by all six models, while 1,522 (21.0%) were missed by at least five models. These consistently misclassified questions may involve challenges in retrieving or applying specialized operational knowledge, plant-specific configurations, technical procedures, and regulatory requirements. Furthermore, 1,337 questions were answered incorrectly by all four lower-performing models but correctly by at least one of the two frontier models (Grok 4.6 or Claude Opus 5.5). This distinction suggests that a subset of challenging questions may be addressed through more capable models and improved reasoning capabilities. In contrast, questions that remain unanswered across all evaluated models may require targeted domain adaptation or retrieval-augmented generation using authoritative nuclear engineering documentation to improve access to specialized domain knowledge.

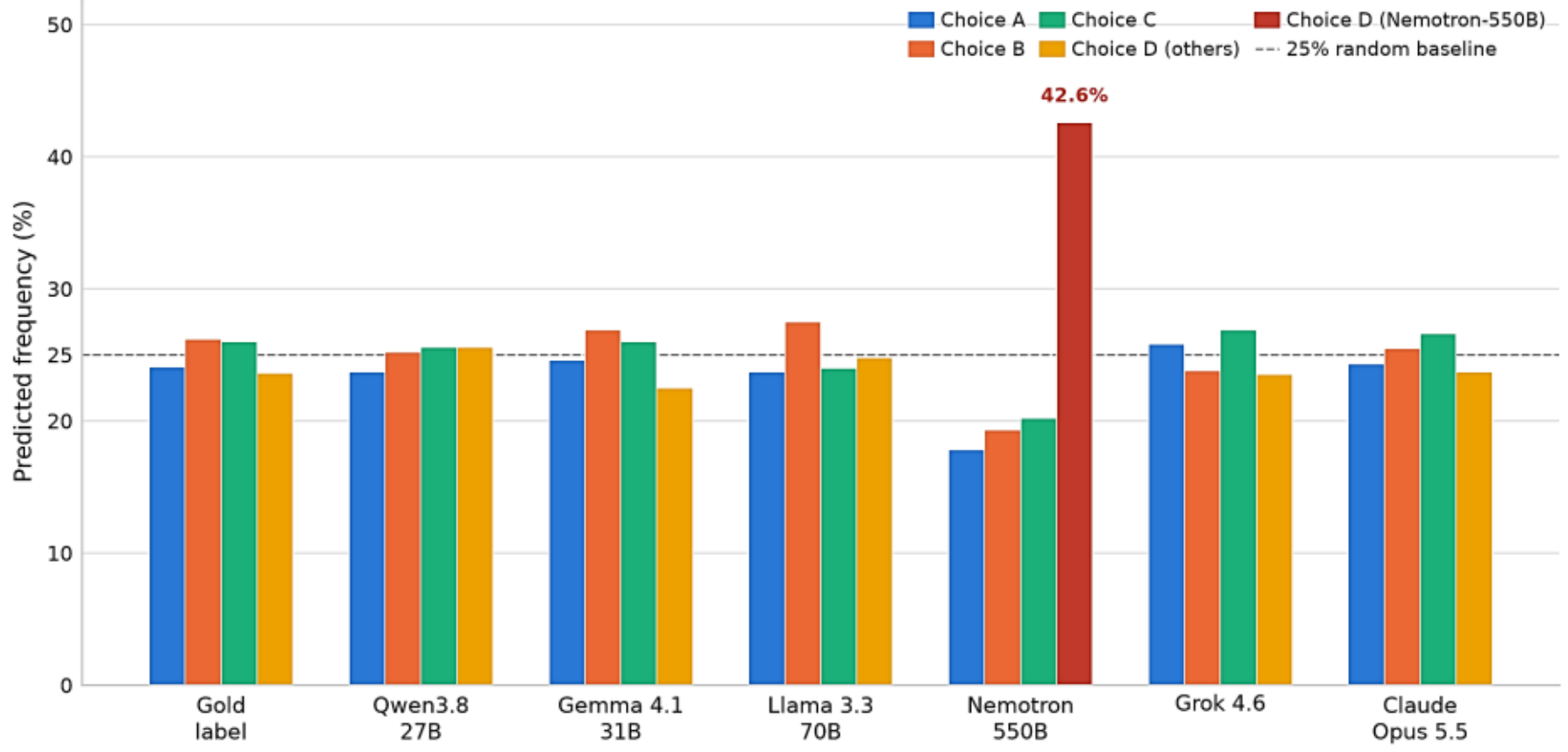

**Fig. 2.** Predicted answer-choice distributions for each model

## 4. Model Reasoning Analysis

### 4.1 Judge Model and Rubric Design

Each model response was evaluated by GPT-5.6-Terra, a frontier LLM used as an automated judge. For each question, the judge received the question, four answer choices, ground truth explanation, and complete model response, including any visible reasoning. A five-dimensional rubric was developed by a nuclear engineering expert to assess the quality of model explanations. A preliminary binary criterion (D1) identified responses with correct final answers, and only these responses were included in the subsequent rubric-based evaluation. Responses with incorrect final answers were excluded to distinguish the quality of successful responses from overall answer accuracy, specifically characterizing how the model reached the correct answers. Each rubric was rated on a five-point integer scale (1–5). To assess the reliability of automated scoring, a domain expert independently reviewed a sample of 55 model responses. The judge and the expert reached the same scoring conclusions on this sample set, confirming adequate inter-rater reliability for the automated evaluation.

- **D2** assessed understanding of the underlying concepts through three subdimensions: **D2a (Problem Comprehension)** evaluated whether the response correctly identified the physical or operational scenario described in the question; **D2b (Correct Theory/Operational Understanding)** assessed the application of appropriate nuclear engineering principles, physical laws, and operational procedures; **D2c (Correct Facts/Procedure)** evaluated the accuracy of specific factual claims, system parameters, and procedural steps.
- **D3 (Distractor Analysis)** assessed the quality of explanations for why the incorrect answer choices were unsuitable.
- **D4 (Clarity and Relevance)** evaluated the organization, relevance, and directness of the response.
- **D5 (Technical Precision)** assessed the appropriate use of nuclear engineering terminology and the precision of quantitative statements.

### 4.2 Model Reasoning Patterns

Table 3 reports the mean rubric scores across correctly answered questions for all six models. Three key findings emerge.

Table 3. Mean rubric scores for each model across correctly predicted responses (1–5 scale).

| Model | N(correct) | D2a | D2b | D2c | D3 | D4 | D5 |
|---|---|---|---|---|---|---|---|
| Qwen3.8-27B | 2,531 | 4.33 | **3.27** | 2.95 | 3.14 | 4.55 | 3.18 |
| Gemma 4.1 | 3,099 | 4.54 | 3.68 | 3.25 | 3.52 | 4.66 | 3.58 |
| Llama 3.3 70B | 2,563 | **4.21** | 3.27 | **2.85** | **2.75** | 4.68 | 3.45 |
| Nemotron-Ultra-550B | 3,278 | 4.77 | 3.49 | 3.11 | 3.32 | 3.74 | 3.57 |
| Grok 4.6 | 5,137 | **4.86** | **4.18** | **3.86** | 3.61 | 4.95 | 4.03 |
| Claude Opus 5.5 | 5,555 | 4.79 | 4.04 | 3.68 | **3.65** | 4.46 | 3.79 |
| Average | | 4.59 | 3.65 | 3.28 | 3.33 | 4.51 | 3.60 |

First, correct answers were not consistently supported by strong technical justifications. Across models, the highest mean scores were observed for problem comprehension (D2a=4.59) and clarity and relevance (D4=4.51), whereas lower scores were obtained for theoretical and operational understanding (D2b=3.65), technical precision (D5=3.60), distractor analysis (D3=3.33), and factual and procedural accuracy (D2c=3.28). For every model, the difference between its highest and lowest dimension scores exceeded one point on the five-point scale, with the lowest score occurring in either D2c or D3. These results suggest that models generally identified the relevant physical or operational scenarios and communicated their responses coherently but were less consistent in providing factually accurate, procedurally sound, and well-supported reasoning.

Second, inter-model differences were more pronounced in reasoning quality than in problem comprehension. D2a scores exhibited a relatively narrow range (4.21–4.86), whereas D2b, D2c, and D3 each spanned approximately 0.9–1.0 points. Grok 4.6 and Claude Opus 5.5 achieved the highest scores on these reasoning-related dimensions, and also produced the most correct answers, suggesting an association between answer accuracy and reasoning quality. Nevertheless, their distractor analysis scores (D3 = 3.61 and 3.65, respectively) remained substantially below their comprehension scores, indicating that systematically explaining why alternative answers are incorrect remains challenging even for frontier models.

Third, response clarity did not necessarily reflect technical quality. Llama 3.3 70B achieved the second-highest clarity score (D4 = 4.68) despite having the lowest scores for factual and procedural accuracy (D2c) and distractor analysis (D3). Conversely, Nemotron-Ultra-550B achieved intermediate scores on content-related dimensions but the lowest clarity score (D4 = 3.74). These contrasting results demonstrate that response clarity and technical correctness represent distinct aspects of model performance and that fluency alone is insufficient to assess nuclear domain competence.

### 4.3 Reasoning Behavior Analysis

Among the six evaluated models, Grok 4.6 and Claude Opus 5.5 generated internal reasoning tokens before producing visible responses. This allows an examination of how inference-time computation relates to answer accuracy (Fig. 3).

Greater reasoning effort was consistently associated with lower accuracy in both frontier models. For Grok 4.6, accuracy declined from 90.3% for responses using fewer than 500 reasoning tokens to 39.0% for those exceeding 20,000 tokens. A similar trend was observed for Claude Opus 5.5, with accuracy decreasing from 88.6% at 1–500 thinking tokens to 50.5% at 1,000–2,000 tokens. Incorrect predictions also involved greater reasoning-token consumption in both models. For Grok 4.6, incorrect responses used approximately 1.7 times as many reasoning tokens as correct responses (mean: 7,822 vs. 4,483), with a similar pattern for Claude Opus 5.5 (780 vs. 563).

These consistent trends suggest that extended reasoning did not necessarily improve predictive performance. Reasoning length is more plausibly interpreted as an indicator of question difficulty than as a direct cause of prediction errors. More challenging questions may require longer deliberation while remaining difficult to solve. Nevertheless, reasoning-token consumption may provide a model-specific signal of answer reliability. For example, Grok 4.6 achieved approximately 83% accuracy with fewer than 2,000 reasoning tokens, compared with approximately 47% for responses exceeding 10,000 tokens. Such signals could complement conventional accuracy metrics by identifying high-effort responses for targeted expert review. For Claude Opus 5.5, brief deliberation appeared beneficial, whereas adaptive thinking did not reliably identify questions that could be answered without deliberation. Specifically, 13.3% of responses (1,002 of 7,546) contained no explicit thinking tokens and achieved 74.4% accuracy, 14.2 percentage points below the 88.6% achieved with 1–500 thinking tokens. This finding suggests that automatically skipping deliberation may forgo potential accuracy gains.

The two models also exhibited substantially different computational requirements without a corresponding accuracy advantage from more extensive reasoning. Grok 4.6 generated reasoning traces of up to approximately 43,000 tokens, and 7.2% of its responses required more than 300 s. In contrast, Claude Opus 5.5 used at most approximately 2,660 thinking tokens and completed 97.3% of responses within 20 s. Despite consuming roughly an order of magnitude more reasoning tokens on average, Grok 4.6 answered fewer questions correctly than Claude Opus 5.5. These results highlight the trade-offs among computational cost, response latency, and predictive accuracy, demonstrating that larger reasoning budgets do not necessarily translate into better performance on NEBench R1. However, cross-model latency comparisons should be interpreted cautiously because serving infrastructure also affects response time. All 7,546 Claude Opus 5.5 responses terminated normally, ruling out output truncation as an explanation for its observed accuracy pattern.

## 5. Discussion

NEBench R1 reveals a pronounced three-tier performance hierarchy among contemporary LLMs, with accuracy ranging from 34% to 73.9%. General-purpose open-weight models only modestly above the random-guessing baseline of 25% for four-choice questions, indicating that instruction-following capabilities alone are insufficient for specialized nuclear engineering tasks. Larger open-weight models achieved moderate improvements, although reasoning activation yielded less than one percentage point of improvement for smaller models, and Nemotron-Ultra-550B remained in the intermediate tier. These findings suggest that neither parameter scaling nor extended reasoning alone ensures domain competence. Frontier reasoning models substantially outperformed open-weight models, demonstrating stronger nuclear knowledge and problem-solving capabilities. Nevertheless, even the best-performing model, Claude Opus 5.5, achieved only 73.9% accuracy, below the approximately 80% NRC passing threshold[16]. This persistent performance gap highlights substantial limitations in current LLMs' ability to address specialized nuclear engineering examination questions.

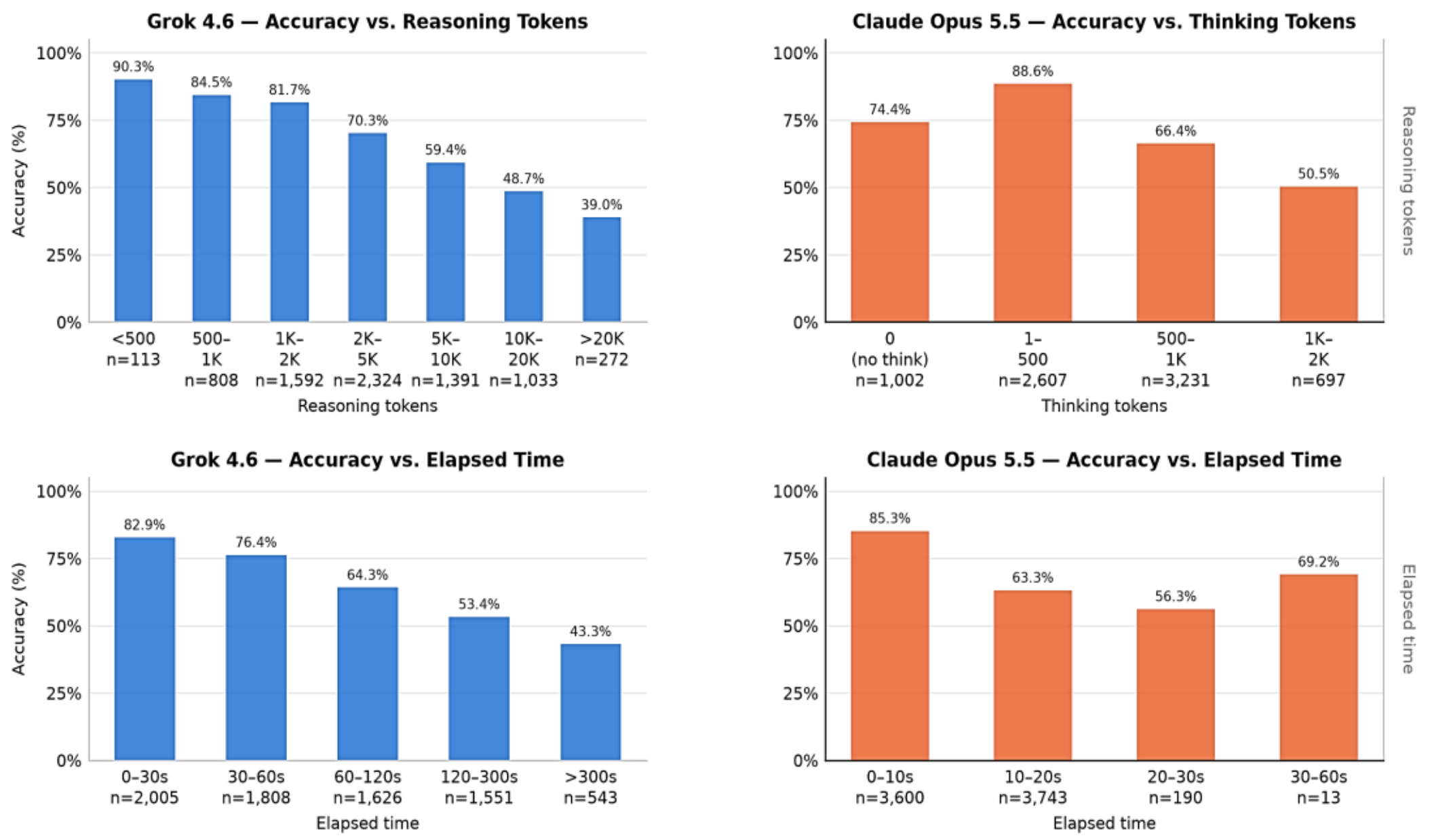

**Fig3.** Accuracy by reasoning token consumption (top) and elapsed inference time (bottom) for Grok 4.6 (left, blue) and Claude Opus 5.5 (right, orange). Sample sizes (n) are shown below each bar label.

Rubric-based evaluation showed consistently high Problem Comprehension (D2a) scores but weaknesses in Correct Theory (D2b) and Correct Facts/Procedure (D2c), indicating that answer accuracy does not necessarily reflect technically sound reasoning. Furthermore, rubric evaluation was restricted to correctly answered questions and therefore characterizes the quality of successful responses rather than the reasoning associated with incorrect predictions. The number of correctly answered questions varied by more than twofold across models (2,531–5,555). Consequently, lower-accuracy models were evaluated on smaller, potentially easier subsets, which may lead to optimistic estimates of their explanatory quality. Finally, the negative association between inference-time computation and accuracy may reflect question difficulty rather than a detrimental effect of extended reasoning, as more challenging questions may require greater computational effort while remaining harder to answer correctly.

Several limitations should be considered when interpreting these results. First, the multiple-choice format permits random guessing (25% expected baseline) and partial answer elimination, without requiring models to independently generate procedural responses or demonstrate complete open-ended reasoning. Second, NRC licensing examination materials are publicly available, and the extent to which these questions may have appeared in model pretraining corpora cannot be quantified without access to training data. Third, automated rubric scoring using GPT-5.6-Terra may introduce systematic evaluation biases. Moreover, restricting rubric evaluation to correctly answered questions leaves the reasoning quality underlying incorrect predictions unexamined.

## 6. Conclusion

This study introduced NEBench R1, a reasoning-oriented benchmark comprising 7,546 unique question-answer pairs derived from U.S. NRC reactor operator licensing examinations across 18 facilities. By incorporating explicit answer explanations and classifications based on the NRC K/A catalogs, the benchmark provides a structured resource for evaluating LLM capabilities in nuclear engineering. Evaluation of six contemporary LLMs demonstrated substantial differences in domain competence, with accuracy ranging from 34% to 73.9%. The pronounced separation between performance tiers demonstrates that NEBench R1 discriminates among levels of LLM nuclear engineering competence, while remaining error rates highlight the difficulty of the benchmark and the limitations of current LLMs for this domain. Rubric-based assessment revealed that correct answer selection does not necessarily imply technically sound reasoning, particularly in the application of nuclear engineering theory, regulatory knowledge, and operational procedures. The results establish a baseline for future research on domain-specific knowledge acquisition, reasoning reliability, and multimodal understanding, while highlighting the need for further improvements before LLMs can support safety-critical decision-making in nuclear engineering.